\documentclass[10pt,letterpaper,twocolumn]{article}
\usepackage[margin=0.75in,columnsep=0.28in]{geometry}
\usepackage[T1]{fontenc}
\usepackage{newtxtext}
\usepackage[hyphens]{url}
\usepackage{graphicx}
\usepackage[round,authoryear]{natbib}
\usepackage{caption}
\usepackage{amsmath}
\usepackage{amssymb}
\usepackage{algorithm}
\usepackage{algorithmic}
\usepackage{newfloat}
\usepackage{listings}
\usepackage{booktabs}
\usepackage{xcolor}
\usepackage{microtype}

\DeclareCaptionStyle{ruled}{labelfont=normalfont,labelsep=colon,strut=off}
\floatstyle{ruled}
\newfloat{listing}{tb}{lst}{}
\floatname{listing}{Listing}

\title{\textbf{IRIS: Reusable Identity Representations from Frozen LLMs for Entity Alignment}}
\author{%
\begin{tabular}{c}
Xinran Liu\textsuperscript{*},
Shengtao Li\textsuperscript{*},
Shouqian Shi\textsuperscript{\dag},
Ge Wang,
Xin-Wei Yao\\[0.35em]
\small \textsuperscript{*}Equal contribution.\quad
\textsuperscript{\dag}Corresponding author: \texttt{sqlite@nju.edu.cn}
\end{tabular}%
}
\date{}

\begin{document}

\maketitle
\begin{abstract}
Entity alignment (EA) identifies entities across knowledge graphs
(KGs) that refer to the same real-world object. Conventional EA
methods mainly exploit explicit graph structures and textual fields,
which often provide insufficient semantic understanding to recognize
the same entity under heterogeneous descriptions and distinguish it
from semantically similar entities. Although large language models
(LLMs) offer deeper entity understanding, existing LLM-based EA
methods largely use this capability for auxiliary generation or
candidate-conditioned decisions. Consequently, such understanding is
not distilled into a stable and directly comparable identity space,
leaving alignment tied to specific KG pairs or candidate sets and
requiring repeated processing as the matching context changes.
To address these limitations, we propose \textbf{IRIS}
(Identity Representations from Internal States), a training-free
framework that constructs for each entity an iris-like signature
encoding its distinctive and stable identity characteristics. IRIS
derives these signatures by eliciting identity-oriented contextual
representations from a frozen LLM, thereby forming a shared space in
which each entity is encoded once and can be aligned across different
KGs through direct similarity comparison, without pair-dependent
representation construction or candidate-wise LLM inference.
Across four established EA benchmarks and two frozen LLM backbones,
the best IRIS variants achieve Hits@1 scores of 100.00, 99.38, 98.31,
and 97.99 on D-Y-15K V2, DBP-WIKI, ICEWS-WIKI, and ICEWS-YAGO,
respectively.
\end{abstract}

\section{Introduction}

Entity alignment (EA) identifies entities across knowledge graphs
(KGs) that refer to the same real-world object and is fundamental to
knowledge integration. Existing methods learn comparable
representations from graph structures, attributes, and textual
descriptions through knowledge graph embeddings
\cite{chen2016multilingual,sun2017cross,sun2018bootstrapping}, graph
neural networks
\cite{wang2018cross,cao2019multi,wu2019relation,sun2020knowledge,
mao2020relational,mao2021boosting}, or multi-view and interaction-based methods
\cite{zhang2019multi,tang2020bert}.

Reliable EA, however, requires more than matching explicit graph and
textual patterns. The same entity may appear under different
languages, abbreviations, naming conventions, or incomplete
descriptions, while different entities may share similar names,
types, and relational contexts. Conventional EA methods often lack
the semantic depth needed to recognize the same entity across
heterogeneous descriptions and distinguish it from semantically
similar entities.

Large language models (LLMs) provide stronger semantic understanding
and have recently been introduced into EA for entity description
generation, information augmentation, and candidate reasoning.
ChatEA performs candidate retrieval followed by LLM-based reasoning
\cite{jiang2024unlocking}; LLM4EA uses LLM-generated annotations
with active learning \cite{chen2024entity}; HLMEA combines candidate
filtering with LLM selection \cite{jin2025hlmea}; and EasyEA
integrates LLM-based summarization into a multi-stage alignment
framework \cite{cheng2025easyea}. These methods demonstrate the value
of LLMs for interpreting heterogeneous entity information.

Nevertheless, existing LLM-based methods mainly use this understanding
for auxiliary generation or candidate-conditioned decisions. The
resulting semantic knowledge is not retained as a stable and directly
comparable identity representation. Consequently, representation
construction or matching remains tied to specific KG pairs,
cross-graph interactions, or candidate sets. Changes in the
counterpart KG or candidate context may therefore require repeated
representation learning, interaction, or LLM inference, limiting both
representation reuse and alignment efficiency.

To address these limitations, we propose \textbf{IRIS}, a
training-free framework that constructs for each entity an iris-like
signature encoding its distinctive and stable identity characteristics.
IRIS directs the contextual representations of a frozen LLM toward
entity identity and converts its semantic understanding into a
reusable vector representation. Like an iris used for identification,
each signature is designed to distinguish one entity from others
while remaining stable across heterogeneous surface realizations.

Each IRIS signature is independently derived from evidence in the
entity's resident KG, without access to a counterpart graph or
candidate set. Entity representations can therefore be computed once,
placed in a shared comparison space, and reused across different KG
combinations. Alignment is subsequently performed through direct
vector similarity, avoiding pair-dependent representation construction
and candidate-wise LLM inference.

The main contributions are summarized as follows:

\begin{itemize}

\item We propose \textbf{IRIS}, an identity-oriented representation
elicitation framework that converts the contextual understanding of a
frozen large language model into stable and discriminative entity
signatures, without entity-alignment supervision or parameter updates.

\item IRIS reformulates entity alignment as independent entity
encoding followed by direct vector comparison. Each entity is encoded
once from its resident knowledge graph and reused across different
graph combinations, avoiding pair-dependent representation
construction and repeated candidate-wise LLM inference.

\item Extensive experiments on D-Y-15K V2, DBP-WIKI, ICEWS-WIKI,
and ICEWS-YAGO demonstrate the effectiveness and generality of IRIS.
The best IRIS variants achieve Hits@1 scores of 100.00, 99.38, 98.31,
and 97.99, respectively, outperforming the strongest compared
LLM-based baseline by 10.31 and 4.49 percentage points on the two
ICEWS benchmarks.

\end{itemize}

\section{Related Work}

\subsection{Entity Representation Learning for Entity Alignment}

Entity representation learning is central to entity alignment. Early
embedding-based methods learn cross-KG comparable representations from
relational triples and seed alignments
\cite{chen2016multilingual,sun2017cross,sun2018bootstrapping}. Graph-based
approaches further aggregate neighborhood and relation information through
graph neural networks
\cite{wang2018cross,cao2019multi,wu2019relation,sun2020knowledge,
mao2020relational,mao2021boosting}. Subsequent methods combine structural
information with entity names and attributes \cite{zhang2019multi}, or
directly model interactions between candidate entity pairs
\cite{tang2020bert}.

These methods progressively enrich entity representations, but their
cross-graph comparability is generally established within a predefined
alignment task through seed supervision, joint optimization, or cross-graph
interaction. IRIS instead constructs an identity representation for each
entity independently from its own KG and performs cross-graph alignment only
after representation construction.

\subsection{LLM-Enhanced Entity Alignment}

Large language models have recently been introduced into EA to provide
semantic knowledge beyond explicit KG evidence. LLM4EA obtains
LLM-generated alignment annotations through active learning and mitigates
annotation noise \cite{chen2024entity}. EasyEA uses LLM-based summarization
together with representation fusion and candidate selection
\cite{cheng2025easyea}. ChatEA retrieves candidate entities and applies
conversational reasoning to refine alignment decisions
\cite{jiang2024unlocking}, while HLMEA combines candidate filtering with
LLM-based selection \cite{jin2025hlmea}.

These methods generally integrate LLM outputs into task-specific alignment
pipelines rather than explicitly constructing reusable identity
representations. In contrast, IRIS directly extracts identity-oriented
contextual representations from a frozen LLM and preserves them as reusable
entity signatures independent of counterpart entities and candidate sets.

\subsection{Decoder-Only LLMs for Representation Learning}

Recent studies have demonstrated that decoder-only LLMs can be repurposed
for representation learning. SGPT develops pooling and fine-tuning
strategies for semantic search with GPT-style models
\cite{muennighoff2022sgpt}. Echo Embeddings improves causal representations
through repeated input construction \cite{springer2402repetition}. LLM2Vec
adapts decoder-only LLMs through attention modification and
representation-oriented training \cite{behnamghader2024llm2vec}, while
NV-Embed develops a specialized general-purpose embedding model
\cite{lee2025nv}.

These methods primarily target general textual similarity. Entity alignment
additionally requires representations that remain stable across
heterogeneous descriptions of the same entity while separating different
entities with similar semantics. IRIS addresses this identity-specific
requirement through structured KG contextualization and semantic readout,
without modifying or optimizing the frozen LLM.

The representation quality of an LLM also varies across Transformer layers.
Previous work has studied intrinsic dimension, anisotropy, and
expansion--compression behavior in Transformer representations
\cite{valeriani2023geometry,razzhigaev2024shape}. IRIS uses intrinsic
dimension as a label-free signal to select a fixed readout layer window for
each backbone.

\section{Method}

\subsection{Problem Setup}

Consider a collection of $K$ knowledge graphs, denoted by $\mathbb{G}=\{G^{(k)}\}_{k=1}^{K}$. Each graph $G^{(k)}=(\mathcal{E}^{(k)},\mathcal{R}^{(k)},\mathcal{T}^{(k)})$ contains a set of entities, relations, and relational triples. Given an entity $e\in\mathcal{E}^{(k)}$, IRIS constructs its identity representation $\mathbf{z}_e$ using only the information associated with $e$ in its resident KG. This construction does not access another KG, cross-graph candidates, or the source--target role of $e$ in a particular alignment task.

Because entity representations constructed from different KGs share a common comparison space, each representation can be retained and reused when the counterpart KG changes.

\subsection{Overview}

\begin{figure*}[t]
    \centering
    \includegraphics[width=\textwidth]{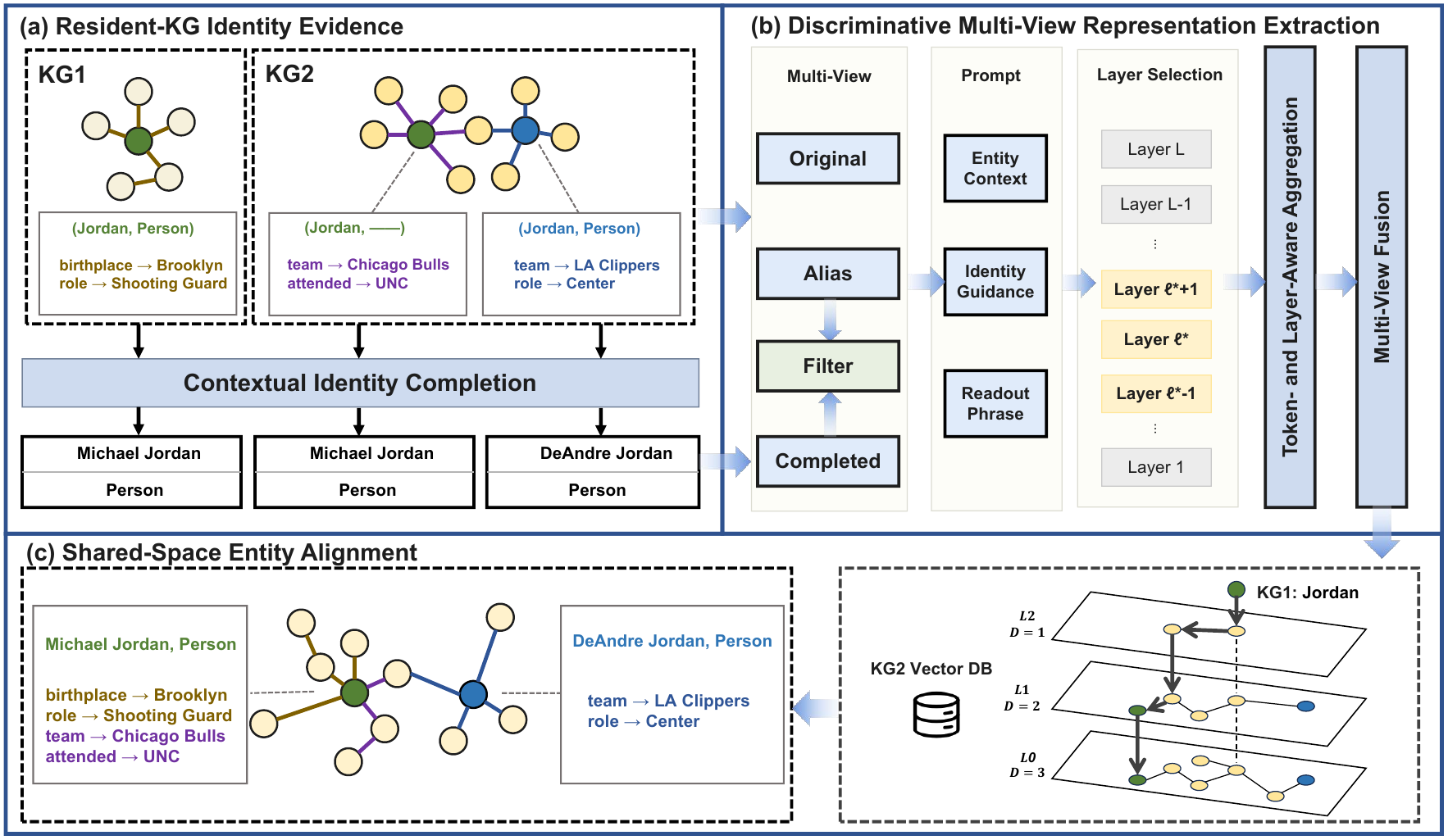}
    \caption{Overview of IRIS. (a) For each entity, IRIS independently
    organizes identity evidence from its resident KG and applies
    contextual identity completion to obtain a complementary full
    English name and entity type. (b) The original name, available
    aliases, and contextually completed name form multiple identity
    views. Each retained view is processed by an identity-oriented
    prompt, followed by intrinsic-dimension-based layer selection and
    token- and layer-aware aggregation. The resulting view
    representations are fused into a reusable entity signature.
    (c) Independently constructed signatures are stored in a shared
    representation space and used for cross-KG entity alignment through
    direct similarity comparison. The vector index illustrates a
    scalable retrieval implementation; all reported experiments use
    exact cosine-similarity ranking.}
    \label{fig:iris_overview}
\end{figure*}

IRIS constructs a reusable identity signature for each entity in
three stages, as illustrated in Figure~\ref{fig:iris_overview}.
First, it organizes the entity and its local KG evidence into a
structured textual context and applies contextual identity completion
to recover a complementary full English name and entity type.
Second, it constructs multiple name views and independently extracts
identity-oriented representations from each retained view using a
frozen LLM. Third, it aggregates the selected token and layer
representations and fuses the resulting view representations into a
unified entity signature. Since each entity is encoded independently
using only evidence from its resident KG, the resulting signatures
can be cached and directly compared across different KGs.

\subsection{Structured Local KG Context Construction}

IRIS organizes the information associated with each entity into a structured textual context, including its name, type, properties, and local relations.

Each entity is represented as \texttt{(\{name\}, \{type\}, \{properties\})}, where \texttt{\{properties\}} denotes the key--value pairs included in the entity record. Together, these fields form a structured description of the entity.

Each outgoing relation is represented as \texttt{this node -[\{relation\}]-> \{target\}}, where \texttt{\{target\}} follows the same entity format. This serialization preserves relation direction and incorporates the neighboring entity's name, type, and properties into the local context. All information is collected solely from the entity's resident KG.

\subsection{Contextual Identity Completion}

The same entity may be described using different languages, incomplete names, or abbreviations. IRIS
therefore uses the structured local context to recover a complementary full English name and type while retaining the original KG information.

Given the serialized entity and its local relations, IRIS adopts the
following generation template:

\begin{center}
\begin{tabular}{@{}p{0.96\columnwidth}@{}}
\hline
\raggedright
\rule{0pt}{2.2ex}%
Focus on entity deduplication and disambiguation.\newline
Entity: \texttt{\{entity\}}\newline
Relationships: \texttt{\{relations\}}\newline
Determine the entity's full English name and type.\newline
\mbox{Answer: \texttt{"}%
\underline{\texttt{[full name]}\vphantom{gj}}%
\texttt{" with type: }%
\underline{\texttt{[type]}\vphantom{gj}}%
\texttt{.}}
\tabularnewline[0.6ex]
\hline
\end{tabular}
\end{center}

Here, \texttt{\{entity\}} denotes the structured description of the
target entity, and \texttt{\{relations\}} contains its serialized local
relations. The underlined fields indicate content generated by the
frozen LLM. IRIS first generates the full English name, then appends the fixed phrase
\texttt{\{" with type:\}} and continues generating the type from the
same KV cache. The completed name and type supplement the original KG fields with
additional identity cues while preserving the information provided by
the KG.

\subsection{Discriminative Multi-View Representation Extraction}

\paragraph{Complementary Name-View Construction.}

IRIS retains the original KG name as the primary view because it is
directly grounded in the source data. Available aliases and the
contextually completed full English name are treated as auxiliary views. These
name variants may express the same entity through translations,
expanded abbreviations, more complete names, or alternative word
orders, thereby providing complementary descriptions of its identity.

To avoid introducing redundant views, IRIS compares each auxiliary
name with the original name using character-level edit distance.
Variants whose distance does not exceed one third of the shorter name
length are removed, and only sufficiently distinct auxiliary names
are retained for subsequent representation extraction.

\paragraph{Identity-Oriented Semantic Readout.}

For each retained name view, IRIS constructs the following
representation prompt:

\begin{center}
\begin{tabular}{@{}p{0.96\columnwidth}@{}}
\hline
\raggedright
Focus on entity deduplication and disambiguation in Knowledge Graph.
The entity: \texttt{\{entity\}}.\newline
Context: \texttt{\{relations\}}\newline
What is the core semantics of the entity
\texttt{\{entity\}}?\newline
The core semantics of entity \texttt{"\{label\}"} \texttt{(\{type\})}
\tabularnewline
\hline
\end{tabular}
\end{center}

The \texttt{\{label\}} field is instantiated with the current name
view, while \texttt{\{type\}} contains the corresponding entity type.
The final label--type expression follows the complete entity context.
Under causal attention, its hidden states incorporate the preceding
entity description and local relations while remaining explicitly
anchored to the identity being represented.

IRIS extracts the post-MLP hidden states corresponding to the label
and type tokens at the outputs of the selected Transformer blocks.
Quotation marks and other fixed delimiters are excluded from the
readout. This design concentrates the representation on an explicit
entity-specific semantic target.

\paragraph{Token- and Layer-Aware Aggregation.}

The label and type tokens contribute differently to entity
representation. IRIS therefore applies a progressive
position-weighting scheme to the label tokens. Regular label tokens
receive linearly increasing weights according to their relative
positions, allowing the complete name to be preserved while reducing
the dominance of generic components appearing earlier in the
sequence. Label tokens containing special symbols are assigned a
smaller fixed weight to suppress formatting-related noise, while type
tokens receive another fixed weight because they provide complementary
categorical information.

Let $\mathcal{P}_{e,v}^{n}$, $\mathcal{P}_{e,v}^{s}$, and
$\mathcal{P}_{e,v}^{t}$ denote the regular label tokens,
special-symbol label tokens, and type tokens, respectively. Suppose
that the regular label contains $m$ tokens. For each selected token
position $j$, let $r_j\in\{0,\ldots,m-1\}$ denote its zero-based
position among the regular label tokens when
$j\in\mathcal{P}_{e,v}^{n}$. Its weight is defined as

\begin{equation}
w_j=
\begin{cases}
w_0+\alpha\frac{r_j}{m}, & j\in\mathcal{P}_{e,v}^{n},\\
w_{\mathrm{sym}}, & j\in\mathcal{P}_{e,v}^{s},\\
w_{\mathrm{type}}, & j\in\mathcal{P}_{e,v}^{t}.
\end{cases}
\end{equation}

Here, $w_0$ is the base weight assigned to the first regular label
token, $\alpha$ controls the progressive increase across the label,
and $w_{\mathrm{sym}}$ and $w_{\mathrm{type}}$ are the fixed weights
assigned to special-symbol and type tokens.

The quality of the extracted representations also varies across
Transformer layers. IRIS selects the readout depth using a label-free
intrinsic-dimension criterion. For each frozen backbone, it estimates
a layer-wise intrinsic-dimension trajectory once on an unlabeled
calibration corpus using a trimmed TwoNN estimator
\cite{facco2017estimating}. Following the expansion--compression
pattern observed in Transformer representations
\cite{valeriani2023geometry,razzhigaev2024shape}, IRIS selects the
first local minimum after the initial expansion and aggregates a small
window of neighboring layers. The selected layer window is fixed for
each backbone and reused across all EA datasets.

Let
$\mathcal{P}_{e,v}=\mathcal{P}_{e,v}^{n}\cup
\mathcal{P}_{e,v}^{s}\cup\mathcal{P}_{e,v}^{t}$
denote all selected label and type token positions, and let
$\mathcal{L}$ denote the selected layer window. The representation of
entity $e$ under name view $v$ is computed as

\begin{equation}
\mathbf{z}_{e,v}
=
\operatorname{Norm}
\left(
\frac{1}{|\mathcal{L}|}
\sum_{\ell\in\mathcal{L}}
\frac{\sum_{j\in\mathcal{P}_{e,v}}w_j\mathbf{h}_{j}^{\ell}}
{\sum_{j\in\mathcal{P}_{e,v}}w_j}
\right),
\end{equation}

where $\mathbf{h}_{j}^{\ell}$ denotes the post-MLP hidden state of
token $j$ at the output of Transformer layer $\ell$, and
$\operatorname{Norm}(\cdot)$ denotes $\ell_2$ normalization. The
concrete token weights and selected layer windows are provided in the
implementation details.

\paragraph{Multi-View Representation Fusion.}

The original KG name provides a representation directly grounded in
the entity record, while the retained auxiliary names contribute
complementary name information. IRIS first averages the
representations of all retained auxiliary views and then combines the
result with the original-name representation:

\begin{equation}
\mathbf{z}_e
=
\operatorname{Norm}
\left(
(1-\lambda)\mathbf{z}_{e,o}
+
\frac{\lambda}{|\mathcal{V}_{e}^{+}|}
\sum_{v\in\mathcal{V}_{e}^{+}}
\mathbf{z}_{e,v}
\right),
\end{equation}

where $\mathbf{z}_{e,o}$ denotes the representation extracted from
the original KG name, $\mathcal{V}_{e}^{+}$ is the set of retained
auxiliary name views, and $\lambda$ controls their contribution. When
no auxiliary view is retained, IRIS directly uses
$\mathbf{z}_{e,o}$.

\subsection{Unified Similarity-Based Alignment}

Each IRIS representation is constructed independently using only information from the entity's resident KG. Representations from different KGs can therefore be compared directly in a shared representation space. Since all entity representations are $\ell_2$-normalized, the alignment score between entities $e_i$ and $e_j$ is computed as

\begin{equation}
s(e_i,e_j)=\mathbf{z}_{e_i}^{\top}\mathbf{z}_{e_j}.
\end{equation}

For standard EA evaluation, the similarity scores between the two entity sets are used to rank candidate counterparts. Representation construction requires no cross-graph retrieval, pair-dependent prompt construction, or candidate-wise LLM inference. Once an entity has been encoded, its representation can be cached and reused when aligning the resident KG with other KGs.

\section{Experiments}

\begin{table*}[t]
\centering
\scriptsize
\setlength{\tabcolsep}{2pt}
\begin{tabular}{l ccc ccc ccc ccc}
\toprule
\textbf{Method}
& \multicolumn{3}{c}{\textbf{D-Y-15K V2}}
& \multicolumn{3}{c}{\textbf{DBP-WIKI}}
& \multicolumn{3}{c}{\textbf{ICEWS-WIKI}}
& \multicolumn{3}{c}{\textbf{ICEWS-YAGO}} \\
\cmidrule(lr){2-4}
\cmidrule(lr){5-7}
\cmidrule(lr){8-10}
\cmidrule(lr){11-13}
& Hits@1 & Hits@10 & MRR
& Hits@1 & Hits@10 & MRR
& Hits@1 & Hits@10 & MRR
& Hits@1 & Hits@10 & MRR \\
\midrule

AlignE
& 86.4 & 97.0 & 90.2
& 56.6 & 82.7 & 65.5
& 5.7 & 26.1 & 12.2
& 1.9 & 11.8 & 5.5 \\

BootEA
& 95.0 & 98.6 & 96.3
& 74.8 & 89.8 & 80.1
& 7.2 & 27.5 & 13.9
& 2.0 & 12.0 & 5.6 \\

GCN-Align
& 82.6 & 94.9 & 87.2
& 49.4 & 75.6 & 59.0
& 4.6 & 18.4 & 9.3
& 1.7 & 8.5 & 3.8 \\

RDGCN
& 85.4 & 93.2 & 88.3
& 97.4 & 99.4 & 98.0
& 6.4 & 20.2 & 9.6
& 2.9 & 9.7 & 4.2 \\

Dual-AMN
& 97.5 & 99.3 & 98.1
& 98.3 & 99.6 & 99.1
& 8.3 & 28.1 & 14.5
& 3.1 & 14.4 & 6.8 \\

LLM4EA
& \underline{97.9} & \underline{99.6} & \underline{98.5}
& -- & -- & --
& -- & -- & --
& -- & -- & -- \\

BERT-INT
& -- & -- & --
& \textbf{99.6} & 99.7 & \underline{99.6}
& 56.1 & 70.0 & 60.7
& 75.6 & 85.9 & 79.3 \\

Simple-HHEA
& -- & -- & --
& 97.5 & 99.1 & 98.8
& 72.0 & 87.2 & 75.4
& 84.7 & 91.5 & 87.0 \\

ChatEA
& -- & -- & --
& \underline{99.5} & \textbf{100.0} & \textbf{99.8}
& \underline{88.0} & 94.5 & 91.2
& 93.5 & \underline{95.5} & 94.4 \\

\midrule

IRIS-Llama~3.1-8B
& \textbf{100.00} & \textbf{100.00} & \textbf{100.00}
& 99.38 & 99.89 & \underline{99.60}
& \textbf{98.31} & \underline{99.26} & \textbf{98.70}
& \underline{97.92} & \textbf{99.05} & \underline{98.34} \\

IRIS-Qwen~2.5-7B
& \textbf{100.00} & \textbf{100.00} & \textbf{100.00}
& 99.34 & \underline{99.93} & 99.59
& \textbf{98.31} & \textbf{99.36} & \underline{98.67}
& \textbf{97.99} & \textbf{99.05} & \textbf{98.37} \\

\bottomrule
\end{tabular}
\caption{Main results on four entity alignment benchmarks. Scores are
percentages. Baseline results on D-Y-15K V2 are taken from the
unified GPT-4 annotation setting reported by LLM4EA
\cite{chen2024entity}, while baseline results on DBP-WIKI,
ICEWS-WIKI, and ICEWS-YAGO are taken from the unified evaluation
reported by ChatEA \cite{jiang2024unlocking}. Boldface and
underlining denote the best and second-best results, respectively.
Results tied at the same rank receive the same formatting. Baseline
scores retain the numerical precision reported in their corresponding
sources, while IRIS results are reported with two decimal places.
A dash indicates that the corresponding result was not reported under
the same benchmark setting.}
\label{tb:main_result}
\end{table*}

\subsection{Experimental Setup}

\paragraph{Datasets.}

We evaluate IRIS on four established entity alignment benchmarks:
D-Y-15K V2, DBP-WIKI, ICEWS-WIKI, and ICEWS-YAGO. D-Y-15K V2 is
the dense OpenEA benchmark constructed from DBpedia and YAGO. We use
the same dataset version and evaluation setting as LLM4EA
\cite{chen2024entity}. DBP-WIKI aligns entities between DBpedia and
Wikidata, while ICEWS-WIKI and ICEWS-YAGO align event-oriented
entities from ICEWS with Wikidata and YAGO, respectively. For these
three benchmarks, we follow the dataset versions and evaluation
settings used by ChatEA \cite{jiang2024unlocking}. The two ICEWS
benchmarks exhibit substantially greater structural and descriptive
heterogeneity than DBP-WIKI.

\paragraph{Evaluation Metrics.}

We report Hits@1, Hits@10, and mean reciprocal rank (MRR), expressed
as percentages. Hits@$k$ measures the proportion of source entities
whose correct counterparts appear among the top-$k$ ranked target
entities. MRR averages the reciprocal rank of the correct counterpart
and evaluates the overall ranking quality. Since Hits@1 directly
measures whether the correct counterpart is ranked first, we use it
as the primary metric in the ablation studies.

\paragraph{Baselines.}

We compare IRIS with representative embedding-based, graph-based,
interaction-based, and LLM-enhanced EA methods. The compared methods
include AlignE \cite{sun2020benchmarking}, BootEA
\cite{sun2018bootstrapping}, GCN-Align
\cite{wang2018cross}, RDGCN \cite{wu2019relation}, Dual-AMN
\cite{mao2021boosting}, BERT-INT \cite{tang2020bert}, Simple-HHEA
\cite{jiang2024practical}, LLM4EA \cite{chen2024entity}, and ChatEA
\cite{jiang2024unlocking}.

For D-Y-15K V2, baseline results are taken from the unified GPT-4
annotation setting reported by LLM4EA \cite{chen2024entity}. For
DBP-WIKI, ICEWS-WIKI, and ICEWS-YAGO, baseline results are taken from
the unified evaluation reported by ChatEA
\cite{jiang2024unlocking}. Results unavailable under the
corresponding benchmark setting are marked with a dash.
\paragraph{Implementation Details.}

For the main experiments, we instantiate IRIS with
Llama~3.1-8B-Instruct \cite{grattafiori2024llama3} and
Qwen~2.5-7B-Instruct \cite{yang2024qwen25} as frozen decoder-only
backbones. For brevity, they are denoted as Llama~3.1-8B and
Qwen~2.5-7B, respectively, in all tables. No entity alignment labels
are used to update the model parameters. To examine whether the
effects of individual components generalize across model families and
scales, the ablation studies additionally include Qwen~3.5-4B
\cite{qwen2026qwen35}.

For each entity, IRIS organizes its name, type, properties, and
outgoing relations using the formats described in the method section.
The contextual identity completion stage generates a complementary
full English name and entity type from the local KG context. The
original KG name is always retained as the primary name view, while
available aliases and the contextually completed full English name are treated
as auxiliary views.

To avoid repeatedly encoding nearly identical name forms, IRIS
computes the character-level Levenshtein distance $d(a,b)$ between
the original name $a$ and each auxiliary name $b$. The auxiliary name
is retained only when
$d(a,b)>\lfloor\min(|a|,|b|)/3\rfloor$; otherwise, it is removed as
a near-duplicate. The same criterion is used across all datasets and
backbones.

For each retained name view, IRIS extracts post-MLP hidden states from
the label and type tokens at the outputs of the selected Transformer
blocks. Quotation marks and other fixed delimiters are excluded from
the readout. Let the regular label tokens be indexed by
$i=0,\ldots,m-1$. Their weights are defined as

\begin{equation}
w_i^{\mathrm{label}}
=
1.0
+
0.5\frac{i}{m}.
\end{equation}

A label token containing special symbols is assigned a weight of
$0.25$, while each type token receives a weight of $0.75$. The
selected token representations are then combined through weighted
averaging.

For each backbone, IRIS computes its layer-wise intrinsic-dimension
trajectory once using unlabeled WikiText-103 sequences
\cite{merity2017pointer}. At each Transformer layer, key-projection
states are mean-pooled over valid token positions to obtain sequence
representations, and their intrinsic dimension is estimated using a
trimmed TwoNN estimator \cite{facco2017estimating}. The key-projection
states are used only for label-free layer selection, whereas the final
entity representations are extracted from the post-MLP hidden states
at the selected layers.

We compare four layer windows: the final layers ($L_1$), the global
minimum-ID region ($L_2$), the global maximum-ID region ($L_3$), and
the first local minimum after the initial ID expansion ($L_4$). Using
one-based Transformer-layer indices, the respective windows are
$[30,31,32]$, $[24,25,26]$, $[12,13,14]$, and $[1,2,3,4]$ for
Llama~3.1-8B; $[26,27,28]$, $[24,25,26]$, $[18,19,20]$, and
$[2,3,4,5]$ for Qwen~2.5-7B; and $[30,31,32]$, $[1,2,3]$,
$[18,19,20]$, and $[3,4,5]$ for Qwen~3.5-4B. IRIS uses $L_4$,
whose window is fixed for each backbone and reused across all EA
datasets.

When multiple auxiliary name views are retained, their normalized
representations are first averaged. IRIS then combines the resulting
auxiliary representation with the original-name representation using
equal weights:

\begin{equation}
\mathbf{z}_e
=
\operatorname{Norm}
\left(
\frac{1}{2}\mathbf{z}_{e,o}
+
\frac{1}{2}\mathbf{z}_{e,a}
\right),
\end{equation}

where $\mathbf{z}_{e,o}$ denotes the original-name representation and
$\mathbf{z}_{e,a}$ denotes the average representation of the retained
auxiliary views. When no auxiliary view is retained,
$\mathbf{z}_{e,o}$ is used directly.

All final entity representations are $\ell_2$-normalized. For each
benchmark, we compute the complete cross-KG similarity matrix using
cosine similarity and rank all target entities for each source
entity. IRIS does not use a separate candidate-generation stage,
reranking, pair-dependent prompts, or candidate-conditioned LLM
inference.
\subsection{Main Results}

Table~\ref{tb:main_result} reports the main results on the four
benchmarks. IRIS achieves the best Hits@1 performance on D-Y-15K V2,
ICEWS-WIKI, and ICEWS-YAGO, while remaining competitive on the
nearly saturated DBP-WIKI benchmark.

On D-Y-15K V2, both IRIS-Llama~3.1-8B and IRIS-Qwen~2.5-7B achieve
100.00 on Hits@1, Hits@10, and MRR. Compared with LLM4EA, the
strongest reported baseline under the same benchmark setting, IRIS
improves Hits@1 by 2.10 percentage points. This result demonstrates
that independently constructed identity representations can support
highly accurate alignment without entity-alignment supervision or
candidate-conditioned LLM inference.

The advantages of IRIS are particularly pronounced on the more
heterogeneous ICEWS benchmarks. On ICEWS-WIKI, both IRIS variants
achieve 98.31 Hits@1, outperforming ChatEA by 10.31 percentage
points. IRIS-Qwen~2.5-7B reaches 99.36 Hits@10, while
IRIS-Llama~3.1-8B obtains the highest MRR of 98.70. On ICEWS-YAGO,
IRIS-Qwen~2.5-7B achieves 97.99 Hits@1 and 98.37 MRR, exceeding
ChatEA by 4.49 and 3.97 percentage points, respectively.

On DBP-WIKI, IRIS-Llama~3.1-8B obtains 99.38 Hits@1 and 99.60 MRR,
while IRIS-Qwen~2.5-7B obtains 99.34 Hits@1 and 99.93 Hits@10.
Although this benchmark is already nearly saturated, IRIS remains
within 0.22 percentage points of the strongest Hits@1 result while
avoiding alignment-specific parameter training and candidate-wise
LLM reasoning.

The two frozen backbones exhibit highly consistent performance. Their
Hits@1 scores differ by at most 0.07 percentage points across the four
benchmarks. This consistency indicates that the effectiveness of IRIS
is not restricted to a particular frozen LLM backbone.

\subsection{Ablation Studies}

We conduct controlled ablations to examine three components of IRIS:
token-aware weighting, contextual identity completion, and
intrinsic-dimension-based readout-layer selection. The experiments
use Llama~3.1-8B, Qwen~2.5-7B, and Qwen~3.5-4B. All ablation studies
report Hits@1 to provide a compact and consistent comparison. Within
each ablation, all settings other than the evaluated component are
held fixed. For each backbone, the ``With'' columns and $L_4$
correspond to the same complete IRIS configuration. Consequently, the
results for Llama~3.1-8B and Qwen~2.5-7B reproduce the Hits@1 scores
reported in Table~\ref{tb:main_result}.

\paragraph{Effects of Token-Aware Weighting and Contextual Identity
Completion.}

For token-aware weighting, we replace the weighted aggregation in
IRIS with uniform averaging over the same label and type tokens. For
contextual identity completion, we remove the generated full English
name and entity type, retaining only the original KG information and
available aliases. The two column pairs in
Table~\ref{tab:component_ablation} are independent controlled
comparisons.

\begin{table}[t]
\centering
\scriptsize
\setlength{\tabcolsep}{1.5pt}
\begin{tabular}{@{}llcccc@{}}
\toprule
\textbf{Dataset}
& \textbf{Backbone}
& \multicolumn{2}{c}{\shortstack{\textbf{Token-Aware}\\
                                  \textbf{Weighting}}}
& \multicolumn{2}{c}{\shortstack{\textbf{Contextual Identity}\\
                                  \textbf{Completion}}} \\
\cmidrule(lr){3-4}
\cmidrule(lr){5-6}
& & \textbf{Without} & \textbf{With}
& \textbf{Without} & \textbf{With} \\
\midrule

D-Y-15K V2
& Llama~3.1-8B
& \textbf{100.00} & \textbf{100.00}
& 99.31 & \textbf{100.00} \\
& Qwen~2.5-7B
& \textbf{100.00} & \textbf{100.00}
& 99.25 & \textbf{100.00} \\
& Qwen~3.5-4B
& \textbf{100.00} & \textbf{100.00}
& 99.27 & \textbf{100.00} \\

\midrule

DBP-WIKI
& Llama~3.1-8B
& 98.55 & \textbf{99.38}
& 98.01 & \textbf{99.38} \\
& Qwen~2.5-7B
& 98.23 & \textbf{99.34}
& 97.93 & \textbf{99.34} \\
& Qwen~3.5-4B
& 97.89 & \textbf{99.03}
& 97.57 & \textbf{99.03} \\

\midrule

ICEWS-WIKI
& Llama~3.1-8B
& 98.03 & \textbf{98.31}
& 97.51 & \textbf{98.31} \\
& Qwen~2.5-7B
& 98.03 & \textbf{98.31}
& 97.47 & \textbf{98.31} \\
& Qwen~3.5-4B
& 97.61 & \textbf{98.09}
& 96.95 & \textbf{98.09} \\

\midrule

ICEWS-YAGO
& Llama~3.1-8B
& 97.34 & \textbf{97.92}
& 97.30 & \textbf{97.92} \\
& Qwen~2.5-7B
& 97.85 & \textbf{97.99}
& 97.24 & \textbf{97.99} \\
& Qwen~3.5-4B
& 96.50 & \textbf{97.76}
& 97.05 & \textbf{97.76} \\

\bottomrule
\end{tabular}
\caption{Component ablations of token-aware weighting and contextual
identity completion. ``Without'' and ``With'' indicate whether the
corresponding component is excluded or included. The two column pairs
represent independent controlled comparisons. Scores are Hits@1
percentages. The better result within each comparison is shown in
bold; ties are both boldfaced.}
\label{tab:component_ablation}
\end{table}

As shown in Table~\ref{tab:component_ablation}, token-aware weighting
preserves the saturated results on D-Y-15K V2 and improves all nine
dataset--backbone combinations on the remaining benchmarks, with
gains of 0.14--1.26 percentage points. The largest gains for the two
main backbones occur on DBP-WIKI, reaching 0.83 and 1.11 points for
Llama~3.1-8B and Qwen~2.5-7B, respectively. These improvements indicate
that uniformly averaging all readout tokens can dilute the contribution
of informative name and type components. By assigning different weights
according to token position and function, IRIS better preserves
entity-specific identity cues while reducing the influence of
formatting-related or less discriminative tokens. The consistent gains
across backbones further suggest that this effect is not tied to a
particular model architecture.

Contextual identity completion improves all twelve
dataset--backbone combinations by 0.62--1.46 points, with the largest
gains of 1.37--1.46 points on DBP-WIKI. By recovering a more explicit
English name and type from the resident-KG context, it supplements the
original KG fields and aliases with complementary identity evidence.
The consistent gains across datasets and backbones indicate that this
additional information is particularly useful when the original KG
fields and available aliases do not fully expose the entity's identity.

\paragraph{Effect of Readout-Layer Selection.}

To evaluate the intrinsic-dimension-based readout-layer selection
strategy, we compare four layer aggregation schemes. $L_1$ uses a
fixed window over the final Transformer layers. $L_2$ and $L_3$ use
windows around the global minimum-ID and global maximum-ID layers,
respectively. $L_4$, which is used by IRIS, selects a window around
the first local minimum after the initial intrinsic-dimension
expansion. The backbone-specific layer indices are provided in the
implementation details.

\begin{table}[t]
\centering
\scriptsize
\setlength{\tabcolsep}{3pt}
\begin{tabular}{@{}llcccc@{}}
\toprule
\textbf{Dataset}
& \textbf{Backbone}
& $L_1$
& $L_2$
& $L_3$
& $L_4$ \\
\midrule

D-Y-15K V2
& Llama~3.1-8B
& 99.81 & 94.27 & 93.27 & \textbf{100.00} \\
& Qwen~2.5-7B
& 99.99 & 99.95 & 97.93 & \textbf{100.00} \\
& Qwen~3.5-4B
& 98.87 & \textbf{100.00} & 92.48 & \textbf{100.00} \\

\midrule

DBP-WIKI
& Llama~3.1-8B
& 98.58 & 91.25 & 90.15 & \textbf{99.38} \\
& Qwen~2.5-7B
& 98.57 & 98.79 & 97.70 & \textbf{99.34} \\
& Qwen~3.5-4B
& 97.78 & 98.99 & 86.70 & \textbf{99.03} \\

\midrule

ICEWS-WIKI
& Llama~3.1-8B
& 96.04 & 88.25 & 84.21 & \textbf{98.31} \\
& Qwen~2.5-7B
& 95.84 & 95.88 & 88.91 & \textbf{98.31} \\
& Qwen~3.5-4B
& 94.62 & 97.95 & 81.86 & \textbf{98.09} \\

\midrule

ICEWS-YAGO
& Llama~3.1-8B
& 96.51 & 85.62 & 74.59 & \textbf{97.92} \\
& Qwen~2.5-7B
& 96.25 & 96.21 & 81.98 & \textbf{97.99} \\
& Qwen~3.5-4B
& 93.40 & 97.70 & 66.87 & \textbf{97.76} \\

\bottomrule
\end{tabular}
\caption{Effect of readout-layer selection. $L_1$ uses a fixed window
over the final Transformer layers; $L_2$ and $L_3$ use windows around
the global minimum-ID and global maximum-ID layers, respectively; and
$L_4$ selects a window around the first local minimum after the
initial intrinsic-dimension expansion. $L_4$ is used by IRIS. Scores
are Hits@1 percentages. The best result for each dataset--backbone
combination is shown in bold; ties are both boldfaced.}
\label{tab:readout_layer_selection}
\end{table}

As shown in Table~\ref{tab:readout_layer_selection}, $L_4$ achieves
the highest or tied-highest Hits@1 across every evaluated
dataset--backbone combination. The alternative strategies exhibit
greater sensitivity to the backbone and dataset. Although $L_2$
performs strongly with Qwen~3.5-4B, it substantially degrades
Llama~3.1-8B on DBP-WIKI and both ICEWS benchmarks. The global
maximum-ID strategy $L_3$ performs particularly poorly on the
heterogeneous ICEWS benchmarks. The final-layer strategy $L_1$
remains competitive in some settings but is consistently outperformed
by $L_4$.

These results support the layer-selection rule introduced in the
method section. The first local minimum after the initial
intrinsic-dimension expansion provides a more stable readout region
than either the final-layer window or the global
intrinsic-dimension extrema.

\section{Conclusion}

This paper presented \textbf{IRIS}, a training-free framework for
constructing reusable entity identity representations from frozen
decoder-only LLMs. IRIS organizes the information available in each
entity's resident KG into a structured context, applies contextual
identity completion to recover complementary name and type cues, and
extracts identity-oriented representations from selected post-MLP
hidden states. It further combines multiple name views through
token-aware aggregation and intrinsic-dimension-based layer selection,
yielding a unified representation without entity-alignment
supervision, parameter updates, or candidate-conditioned inference. By encoding each entity independently, IRIS decouples representation
construction from subsequent cross-KG matching. The resulting vectors
can therefore be cached, directly compared, and reused when the
counterpart KG changes. Experiments on four benchmarks with
Llama~3.1-8B and Qwen~2.5-7B demonstrate consistently strong
performance, with particularly large improvements on the
heterogeneous ICEWS datasets. The ablation results further verify the
individual contributions of contextual identity completion,
token-aware aggregation, and intrinsic-dimension-based readout-layer
selection. Overall, these findings suggest that the contextual
knowledge encoded in frozen LLMs can be elicited into a shared
identity-oriented representation space, offering a practical and
promising approach to accurate, efficient, and reusable entity
alignment across knowledge graphs.
\bibliographystyle{plainnat}
\bibliography{references}

\end{document}